\documentclass[runningheads]{llncs}

\usepackage[T1]{fontenc}

\usepackage{graphicx}

\usepackage[shortlabels]{enumitem}

\usepackage{wrapfig}

\usepackage{hyperref} 
\hypersetup{
  colorlinks   = true, 
  urlcolor     = teal, 
  linkcolor    = teal, 
  citecolor   = teal 
}
\usepackage{cleveref}
\usepackage{cite} 
\usepackage{color}
\usepackage{tikz-cd}

\definecolor{lime}{HTML}{A6CE39}
\DeclareRobustCommand{\orcidicon}{%
    \begin{tikzpicture}
    \draw[lime, fill=lime] (0,0) 
    circle [radius=0.14 ] 
    node[white] {{\fontfamily{qag}\selectfont \tiny ID}};
    \draw[white, fill=white] (-0.0625,0.095) 
    circle [radius=0.008];
    \end{tikzpicture}
    \hspace{-2mm}
}

\begin{document}
\title{Space filling positionality and the Spiroformer}
%
%
\author{M. Maurin\inst{1}\href{https://orcid.org/0009-0003-1731-0734}{\orcidicon} \and M.{\'A}. Evangelista-Alvarado\inst{2}\href{https://orcid.org/0000-0003-0997-703X}{\orcidicon} \and
P. Su\'arez-Serrato\inst{1}\href{https://orcid.org/0000-0002-1138-0921}{\orcidicon}}
\authorrunning{Maurin, Evangelista-Alvarado, and Su\'arez-Serrato}
%
\institute{
Instituto de Matem\'aticas, Universidad Nacional Aut\'onoma de M\'exico (UNAM), Mexico Tenochtitlan \\ \email {mmaurin@ciencias.unam.mx } \email{pablo@im.unam.mx}
\and
Universidad Nacional Rosario Castellanos, Mexico City, Mexico \email{miguel.eva.alv@rcastellanos.cdmx.gob.mx} 
}
 \maketitle              
\begin{abstract}
Transformers excel when dealing with sequential data. 
Generalizing transformer models to geometric domains, such as manifolds, we encounter the problem of not having a well-defined global order. 
We propose a solution with attention heads following a space-filling curve. 
As a first experimental example, we present the Spiroformer, a transformer that follows a polar spiral on the $2$-sphere.

\end{abstract}

section{Introduction}

The attention mechanism and transformer architectures \cite{vaswani2017attention}, while revolutionizing sequence modeling and natural language processing, have primarily been developed and optimized for linearly ordered data.

\begin{wrapfigure}{r}{.4\textwidth}
\vspace{-10pt}
	\centering	\includegraphics[width=\linewidth]{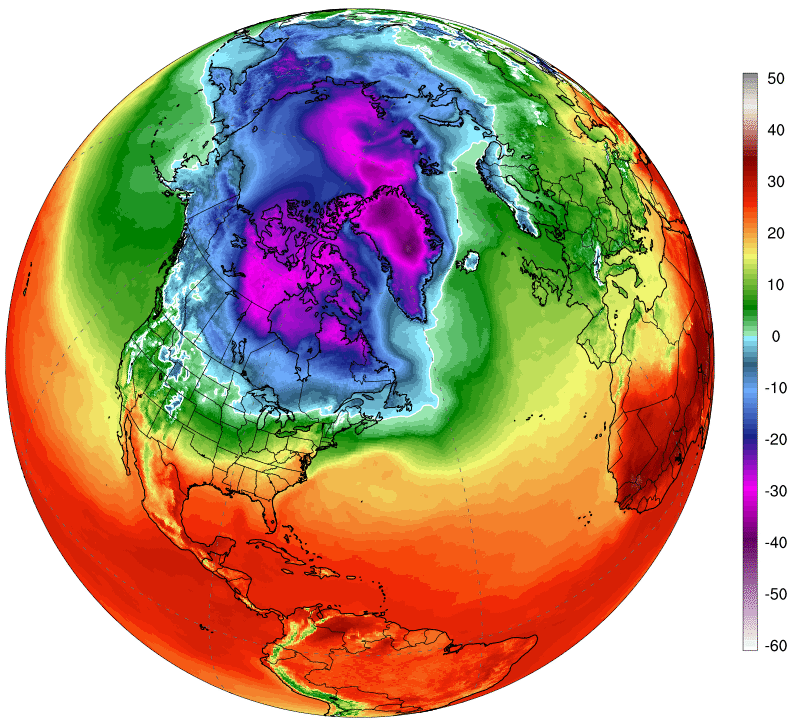}
	\caption{Spherical data may come from geometric domains, such as global environmental sensors (image from ClimateReanalyzer.org).}
        \label{fig:earth-data}
\vspace{-10pt}
\end{wrapfigure}
This focus has led to significant advancements in domains where data inherently possesses a linear or grid-like structure, first on text and now even in images \cite{dosovitskiy2020image}. 

However, the inherent geometry of many real-world datasets, particularly in biological and social networks, often deviates significantly from Euclidean assumptions, demanding a more nuanced geometric approach to capture their underlying complexities.
For example, in \cref{fig:earth-data} we see global temperature data, represented spherically. 

Incorporating geometric information into the positional encoding of transformer architectures presents a promising avenue for extending their capabilities beyond Euclidean domains. 
Traditional positional encodings, designed for linear sequences, fail to capture the intricate relationships and non-Euclidean structures inherent in many datasets.
By embedding geometric information—such as curvature, geodesic distances, or manifold embeddings—directly into the positional encoding, transformers could gain the ability to learn and exploit the underlying geometry of the data, leading to improved performance on tasks involving complex, nonlinear relationships. 
This integration could unlock the potential for transformers to effectively model data from diverse fields such as network biology, social network analysis, and molecular modeling, where geometric structure plays a crucial role.

Recent efforts have focused on adapting transformer architectures to operate effectively on spherical data, addressing the limitations of traditional models designed for Euclidean spaces\cite{cho2022spherical, lai2023spherical, guo2025spatioformer}. 
A spherical self-attention mechanism, that leverages spherical harmonics to respect the rotational symmetries inherent in spherical data has been proposed\cite{russwurm2024spherical}.

 Further advancements introduced equivariance in neural networks for spherical data\cite{cohen2018spherical, ballerin2025so}, emphasizing the importance of equivariance\cite{bogatskiy2022equivariant}, ensuring that model predictions remain consistent under rotations of the input sphere, crucial for applications involving spherical images and other data where rotational invariance is paramount. These works highlight the growing recognition of the need for geometrically aware transformer models in domains where spherical data is prevalent, such as climate modeling and astrophysics.

In this paper, we explore the integration of global manifold structures into transformer positional encodings by employing a manifold-filling curve. 
Specifically, we utilize a spiral trajectory that traverses the 2-sphere, connecting its antipodal poles, to guide the positional encoding, thereby capturing the manifold's inherent geometry and global connectivity beyond local Euclidean approximations.
We find promising experimental results pointing to this being a potentially good idea, as reported in \cref{fig:train-valid}.
Given our current computational resources, the implementations we developed and explain in the following currently exhibit classic overfitting patterns. 
With larger sample sizes on machines with more local memory resources, validation could match training performance close to $90 \%$. 
We open new perspectives on how transformer architecture implementations can take into account the geometric context.

\section{Methods}
The goal of our model is to reconstruct hamiltonian vector fields over the sphere. 
\subsection{Geometric Preliminaries}
\subsubsection{Hamiltonian Vector Fields.}
\begin{figure}[ht]   
    \vspace{-10pt}
    \includegraphics[width=.5\linewidth]{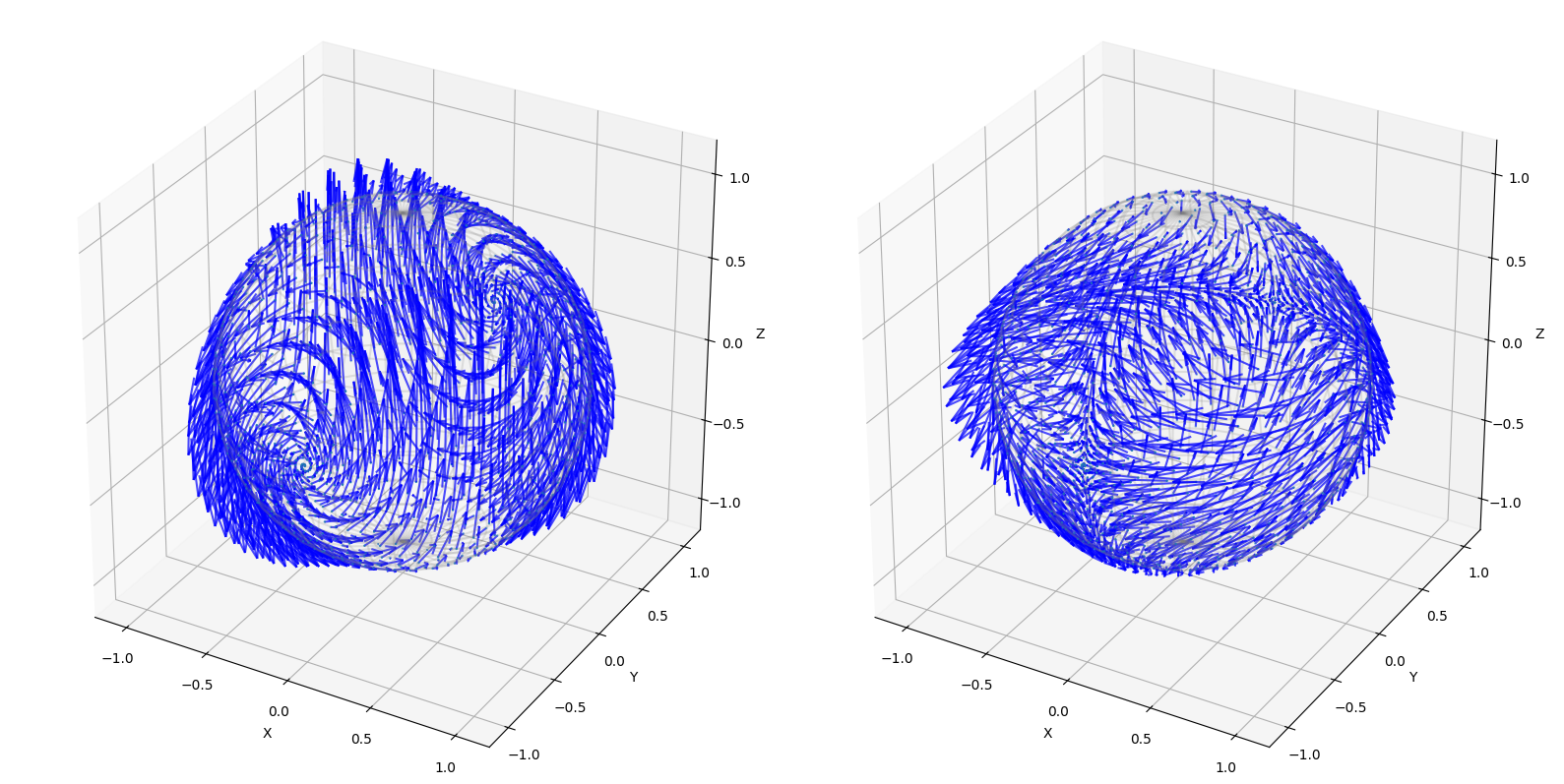} 
    \includegraphics[width=.5\linewidth]{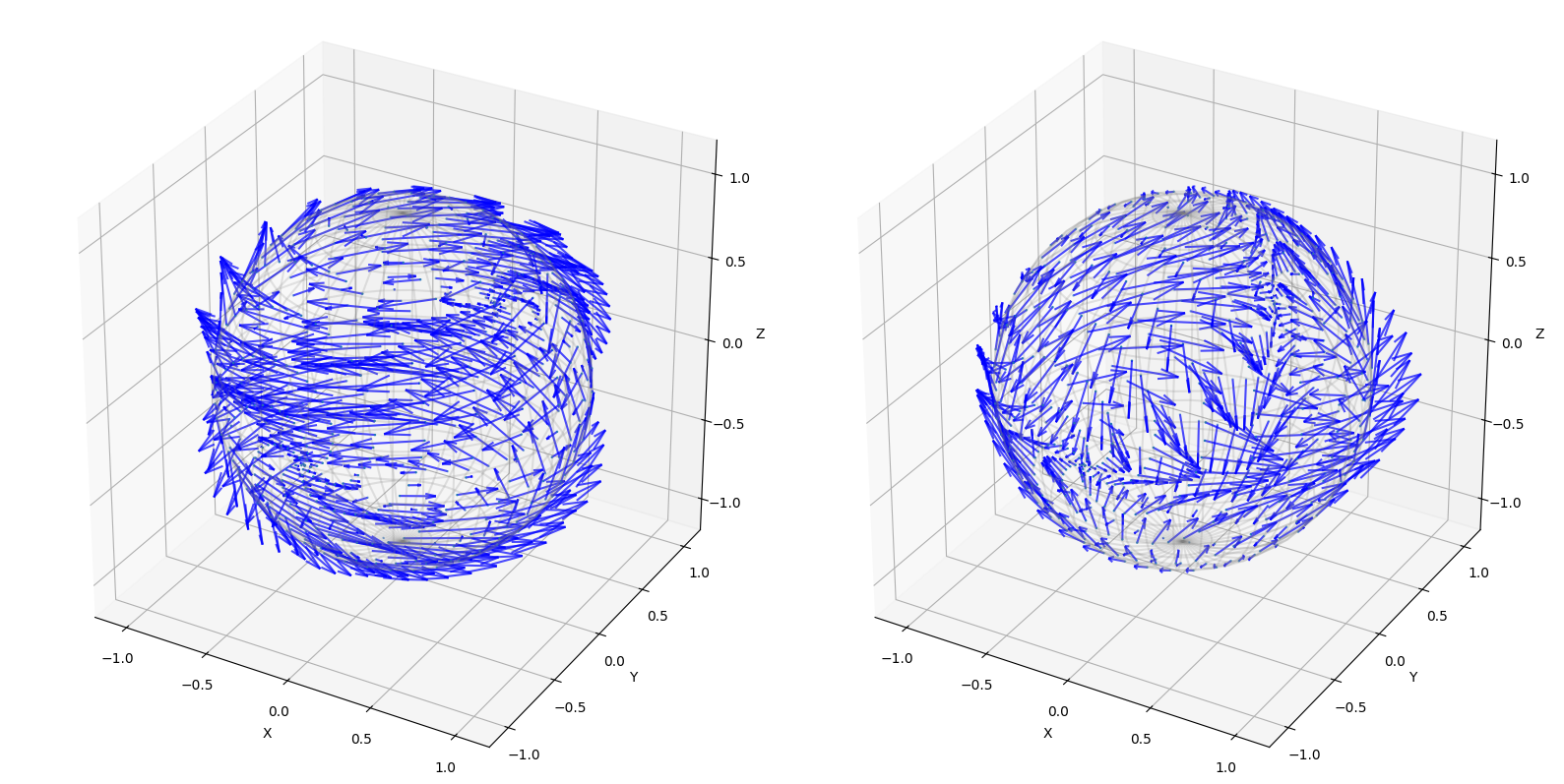} 
    \caption{A selection of spherical Hamiltonian vector fields, showing samples of vectors with base points on a spherical spiral.}
    \label{fig:spherical-Hamiltoninan-vfs}
    \vspace{-10pt}
\end{figure}

 A fundamental concept in mechanics and symplectic geometry, Hamiltonian vector fields arise from functions defined on symplectic manifolds. 
 Recall that a symplectic manifold is a smooth manifold equipped with a closed, non-degenerate 2-form, which allows for the definition of a Poisson bracket and the generation of vector fields from scalar functions\cite{Koszul2019}.
We include some examples of Hamiltonian vector fields on $S^2$ in \cref{fig:spherical-Hamiltoninan-vfs}.
In the specific case of the two-dimensional sphere \(S^2\), we can describe these fields using spherical coordinates \((\theta, \phi)\), where \(\theta\) represents the polar angle and \(\phi\) the azimuthal angle. This representation is particularly useful for modeling physical phenomena that are constrained to spherical geometries, such as fluid flow on a spherical surface, or the dynamics of rotating bodies.

In general, let $(M, \omega)$ be a symplectic manifold, and  $H: M \to \mathbf{R}$ be a smooth function.
There exists a unique vector field on $M$, denoted $X_H$, which is determined by the following equation:
$\iota_{X_{H}}\omega = dH$.
This vector field, $X_H$, corresponds to a dynamical system governed by Hamilton's equations: $\frac{\partial H}{\partial q_{j}}=-{\dot {p_{j}}}$ and $\frac{\partial H}{\partial p_{j}}={\dot {q_{j}}}$, with $j$ ranging from $1$ to $n$. 
The terms $q_j$ stand for position coordinates, while $p_j$ represent the associated momenta; these collectively define the coordinates in the phase space.
The function $H$ is called the Hamiltonian function, and the vector field $X_H$ is its corresponding Hamiltonian vector field.
A system described by the triplet $(M, \omega, H)$ is termed a Hamiltonian system\cite{Arnold1989}. 
Furthermore, the Hamiltonization problem addresses whether a given dynamical system on a smooth manifold can be cast into the Hamiltonian formulation.

In the case of $S^2$ the components of $X_H$ can be expressed in spherical coordinates.
Consider the Poisson bivector $\pi = \sin(\theta) \, \partial_{\theta} \wedge \partial_{\phi}$ on $S^2$.
This bivector allows us to define a vector field from a given smooth function $H$ on $S^2$, known as the Hamiltonian function\cite{CrainicPoisson}. 

The non-degeneracy of the symplectic form $\omega$ implies that it induces an isomorphism of vector bundles $\#_\omega: T^*M \rightarrow TM$ defined by $\omega(\#_\omega(\alpha), Y) = \alpha(Y)$ for any 1-form $\alpha$ and vector field $Y$.
 The inverse of this map, denoted by $b_\omega = (\#_\omega)^{-1}: TM \rightarrow T^*M$, is given by $b_\omega(X) = \omega(X, \cdot)$.
We can use the isomorphism $\#_\omega$ to define a Poisson bivector $\pi$ on $M$. 
A Poisson bivector is a section of the second exterior power of the tangent bundle, $\bigwedge^2 TM$. For any two 1-forms $\alpha, \beta \in \Omega^1(M)$, the Poisson bivector $\pi$ acts as:
        $$\pi(\alpha, \beta) = \omega(\#_\omega(\alpha), \#_\omega(\beta))$$
In local coordinates $(q^i, p_i)$ where the symplectic form has the canonical form $\omega = \sum_i dq^i \wedge dp_i$, the Poisson bivector is given by:
        $$\pi = \sum_i \left( \frac{\partial}{\partial p_i} \wedge \frac{\partial}{\partial q^i} \right) = \sum_i \left( \frac{\partial}{\partial p_i} \otimes \frac{\partial}{\partial q^i} - \frac{\partial}{\partial q^i} \otimes \frac{\partial}{\partial p_i} \right)$$

On a symplectic manifold $(M, \omega)$ with its associated Poisson bivector $\pi$, for any smooth function $H \in C^\infty(M)$ (called the Hamiltonian function), the Hamiltonian vector field $X_H$ is defined by its action on other smooth functions $f \in C^\infty(M)$ through the Poisson bracket $X_H(f) = \{H, f\}$.
 The Poisson bracket $\{H, f\}$ can be expressed in terms of the Poisson bivector $\pi$ and the exterior derivatives of $H$ and $f$, as $\{H, f\} = \pi(dH, df)$.
Using the relationship between the Poisson bivector and the symplectic form, we can rewrite this as:
        $$\{H, f\} = \omega(\#_\omega(dH), \#_\omega(df))$$
The Hamiltonian vector field $X_H$ is related to the Hamiltonian function $H$ through the symplectic form by the equalities 
    $\omega(X_H, Y) = -dH(Y) = -Y(H)$,
        for any vector field $Y$ on $M$.       
        This uniquely defines the Hamiltonian vector field $X_H$, because $\omega$ is non-degenerate.
        In other words, $X_H = \#_\omega(-dH)$.

\begin{wrapfigure}{r}{.4\textwidth}
\vspace{-20pt}
	\centering	\includegraphics[width=0.95\linewidth]{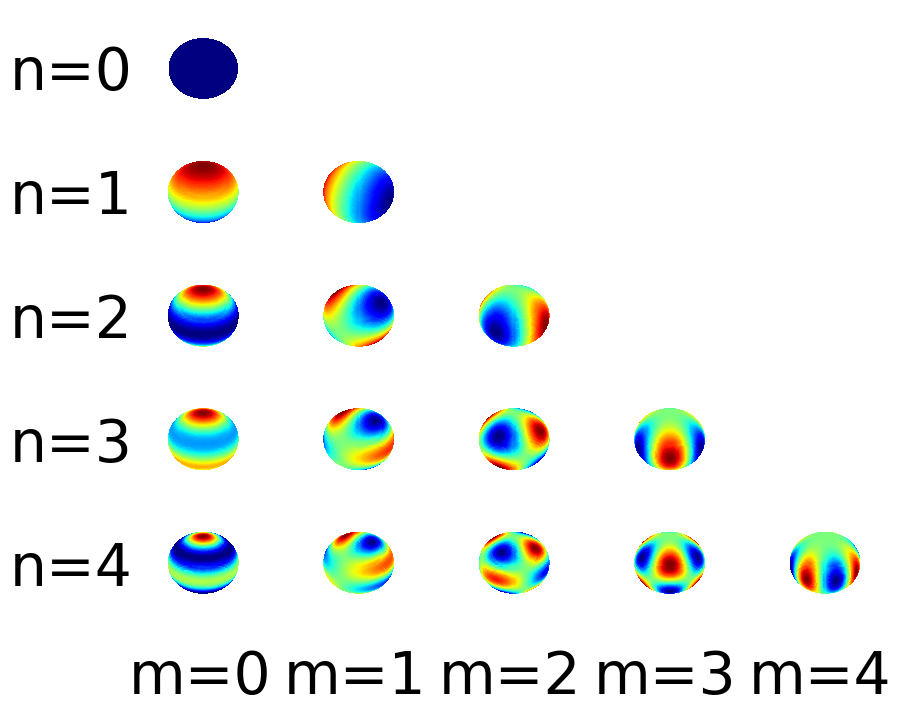}
	\caption{Visualizations of spherical harmonics.}
        \label{fig:spherical-harmonics}
\vspace{-40pt}
\end{wrapfigure}

In many applications, we encounter vector fields that are not inherently Hamiltonian. 
We may seek to approximate such a dynamical system with a Hamiltonian one.
This is known as Hamiltonization. 
Using recent kernel-based methods the underlying Hamiltonian function may be recovered from data to guarantee that the learned model is Hamiltonian \cite{hu2025structurepreserving}.

\subsubsection{Spherical Harmonics}

Spherical harmonics $Y_n^m(\theta, \phi)$ are a set of orthogonal functions on the sphere, indexed by integer degree $n \ge 0$ and order $m$ ($|m|\le l$). 
They form an orthonormal basis often used in physics and are defined using associated Legendre polynomials $P_n^m$ as:
\[
Y_n^m(\theta, \phi) = \sqrt{\frac{(2n+1)}{4\pi}\frac{(n-m)!}{(n+m)!}} P_n^m(\cos\theta) e^{im\phi}
\]
Here, in spherical coordinates, $\theta$ is the polar angle and $\phi$ is the azimuthal.

They form a complete orthonormal basis for the space of square-integrable functions, $L^2$, on the sphere $S^2$. 
So, any function defined on $S^2$ may be expressed as a linear combination of spherical harmonics.
In \cref{fig:spherical-harmonics} we include several examples of spherical harmonics.

\subsubsection{Spiral Over the Sphere.}
To generate an ordered sequence of points on the sphere, we utilize a spherical spiral. This spiral serves as a space-filling curve, providing a continuous traversal of the sphere's surface. The concept of space-filling curves is central to our approach, as it allows us to impose an inherent order on the manifold, enabling the application of sequence-based models like transformers.

The parametric representation of our spherical spiral is given by:
\[
x = \sin(t) \cos(ct), \quad
y = \sin(t) \sin(ct), \quad
z = \cos(t)
\]

where \(t \in [0, \pi]\) is the curve parameter, and \(c\) is a constant that controls the number of turns the spiral makes around the \(z\)-axis. Since we are working on the unit sphere \(S^2\), the radius \(r\) is set to 1.

The spiral's space-filling property ensures that the sampled points cover the sphere in a relatively uniform manner, while maintaining a sequential order. This sequential ordering is crucial for our method, as it allows us to treat the spherical data as a sequence, making it compatible with transformer models that rely on positional embeddings.

\begin{figure}[h!]   
    \centering
    \vspace{-20pt}
    \includegraphics[width=.45\linewidth]{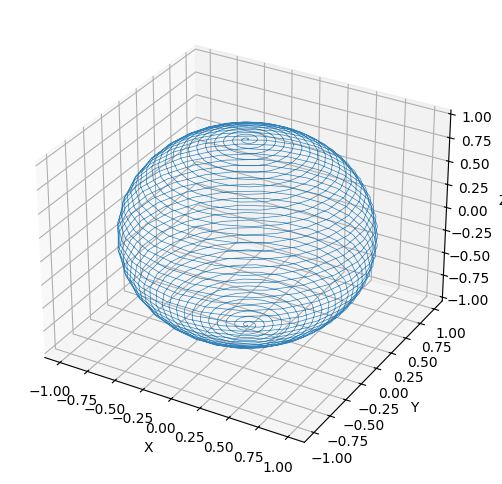}  
    \includegraphics[width=.45\linewidth]{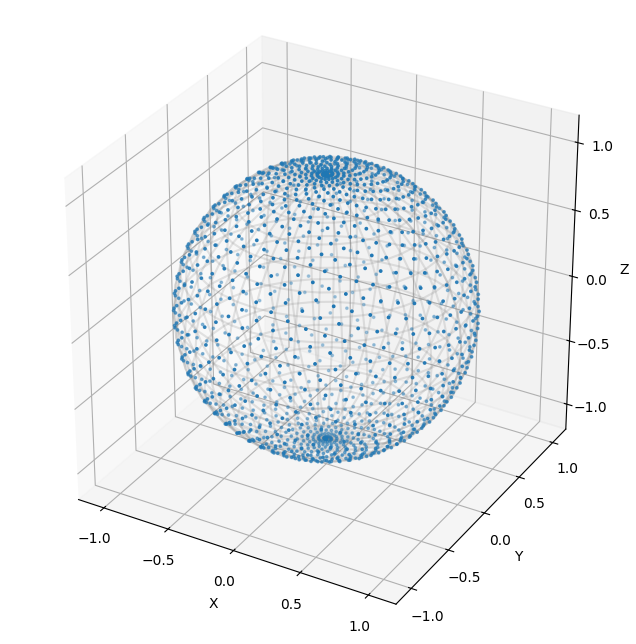}
    \caption{A spherical spiral (left), and a collection of points sampled on it (right).}
    \label{fig:spherical-spiral}
    \vspace{-15pt}
\end{figure}

Using this spherical spiral (see \cref{fig:spherical-spiral}), we transform the problem of processing vector fields on a manifold into a sequence-based task. With this setup, we can now apply a Transformer model over the sphere. 

\subsection{Dataset Generation}

To train a transformer-based model, we require the input data to possess a linear order due to the sequential nature of these models. However, the vector fields defined on the sphere \(S^2\) lack an inherent order.
Therefore, we must define an ordered sampling scheme that allows us to represent the vectors on the sphere in a sequential manner.

We consider the unit sphere \(S^2\) and construct a collection of spherical harmonics that form the basis for defining our Hamiltonian vector fields. Spherical harmonics are defined by two parameters: \(m\) (order) and \(n\) (degree), where \(m\) depends on \(n\). To simplify the problem, we fix a degree of \(n = 32\), which generates a total of \(n^2 = 1024\) spherical harmonics. From each of these spherical harmonics, we compute the corresponding Hamiltonian vector fields.

We utilized the \texttt{sympy} library to generate symbolic expressions for the spherical harmonics. Subsequently, we employed the \texttt{poissongeometry}\cite{poisson} python module to create symbolic expressions of the Hamiltonian vector fields, using the Poisson bivector defined in the previous section.
A guide with examples of how the \texttt{poissongeometry} module works is available \cite{examples}.

To obtain numerical values for the vectors on the sphere, we need a method to sample points in an ordered manner. We use the spherical spiral defined previously, which provides a continuous traversal of the sphere's surface. This spiral ensures that the sampled points cover the sphere relatively uniformly while maintaining a sequential order.

Specifically, we generated a discrete sphere using spherical coordinates with the \texttt{geomstats}\cite{geomstats} library. Then, we used the \texttt{numericalpoissongeometry}\cite{numpoisson} module to numerically evaluate the Hamiltonian vector fields on this discrete sphere. Finally, we sampled the vector fields along the spiral to obtain the sequential data representation.

In summary, the data generation process consists of the following steps:
\begin{enumerate}[A)]
    \item Symbolic Spherical Harmonics:  We generated a list of symbolic spherical harmonics using \texttt{sympy}.
     \item Symbolic Hamiltonian Vector Fields: We created symbolic expressions for the Hamiltonian vector fields using the \texttt{poissongeometry} module and the Poisson bivector $\pi = \sin(\theta) \, \partial_{\theta} \wedge \partial_{\phi}$ mentioned above.
    \item Discrete Sphere Generation: We generated a discrete sphere using spherical coordinates with \texttt{geomstats}.
    \item Numerical Evaluation: We evaluated the Hamiltonian vector fields numerically on the discrete sphere using the \texttt{numericalpoissongeometry} module.
    \item Spiral Sampling: We sampled the vector fields along the spherical spiral to obtain the ordered sequence of vector samples.
\end{enumerate}
This approach provides a structured and ordered dataset that is suitable for training sequence-based models like transformers.

\subsection{Our Model: the {\it Spiroformer}}

Our model leverages the transformer architecture to learn the dynamics of Hamiltonian vector fields on the sphere. The key idea is to transform the vector field reconstruction problem into a sequence-to-sequence learning task, where the input is a sequence of vector field samples along a spherical spiral, and the output is the prediction of the subsequent vector field sample.

We treat segments of the spherical spiral as "sentences", and individual vector field samples along the spiral as "tokens" within those sentences. 
This analogy allows us to directly apply transformer-based models, which have proven highly effective in capturing sequential dependencies.

Given a sequence of vector field samples \(v_1, v_2, ..., v_t\) along the spiral, our model is trained to predict the next sample \(v_{t+1}\). 
By learning such ordered sequences, our model effectively learns to reconstruct Hamiltonian vector fields on the sphere.

\subsubsection{Sequential Representation.} The spherical spiral provides an intrinsic ordering of the vector field samples, allowing us to represent the data as sequences. This sequential representation is essential for applying transformer models.

\subsubsection{Positional Encodings.} To capture the spatial relationships within the vector field, we incorporate positional encodings. These encodings provide the model with information about the location of each vector field sample along the spiral.

\subsubsection{Masking.} During training, we employ masking techniques to prevent the model from "looking ahead" and using future vector field samples to predict the current one. This ensures that the model learns to predict based only on past information, accurately capturing the temporal evolution of the vector field along the spiral.

\begin{figure}
        \vspace{-15pt} 
	\centering	\includegraphics[width=.6\linewidth]{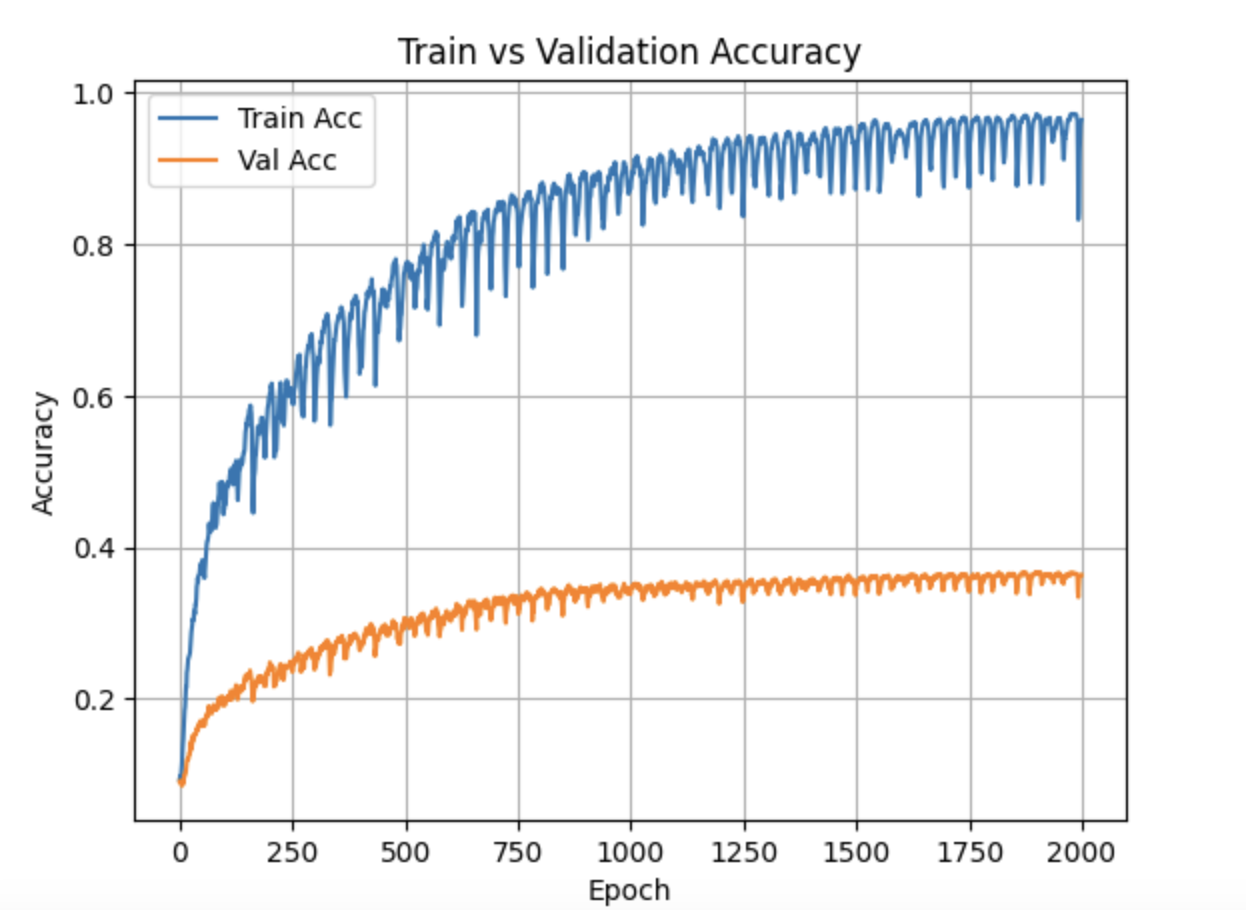}
	\caption{The graph shows our Spiroformer training and validation performances. We used the Optuna optimizer for parameter search, always obtaining similar results. For this particular experiment we used $2$ layers, $4$ attention heads, a dropout rate of $0.2$, and trained for $2000$ epochs.}
    \label{fig:train-valid}
\end{figure}

In essence, our model learns the evolution of the vector field as we traverse the sphere along the spiral.  
As shown in \cref{fig:train-valid}, our model experimentally demonstrates that we can reconstruct the dynamics of spherical Hamiltonian vector fields.
The attained training accuracy confirms the validity of our methods.

\subsection{Results}

Our Spiroformer model achieves high accuracy during training, $\sim 90\%$, as can be seen in \cref{fig:train-valid}. 
These experiments are reported with a database of vector fields, each of which is represented by their evaluation on $100$ points on the sphere.

\section{Conclusions}
This work proposes space-filling curves as a new path to generalize transformers into geometric domains. 
We find that our Spiroformer model learns certain dynamical features and achieves high accuracy during training.
However, the lower validation scores in \cref{fig:train-valid} indicate overfitting. 
In addressing the critical challenge of model generalization and mitigating overfitting, we propose to survey established strategies encompassing regularization methods (such as dropout and weight decay), data augmentation, optimization procedure refinements (like early stopping and learning rate scheduling), and architectural capacity control. 
These techniques collectively could guide the optimization process towards solutions within the parameter space that exhibit lower complexity and improved performance on unseen data.  

A potential path to ensure the model's output is truly Hamiltonian may involve adapting a workflow similar to that of SymFlux \cite{symflux}. 
Thus recovering the symbolic Hamiltonian function directly from vector field data.
Applying this structure-preserving methodology to spherical data is beyond the scope of this paper and remains a compelling direction for future research.

Nevertheless, our contribution is to present a novel method for reading manifold information into a transformer model by following a space-filling curve adapted contextually to intrinsically geometric data.

\begin{credits}
\subsubsection{\ackname} 
PSS thanks the Geometric Intelligence Laboratory at UC Santa Barbara for the welcoming and stimulating environment during fall 2024, as well as the organizers of the AI in Mathematics and Theoretical Computer Science meeting at the Simons Institute for the Theory of Computing at UC Berkeley in spring 2025. 
\subsubsection{\discintname}
The authors have no competing interests to declare that are
relevant to the content of this article. 
\end{credits}
%
%


\begin{thebibliography}{8}

\bibitem{Arnold1989}
V.I. Arnold, A. Weinstein, K. Vogtmann: Mathematical Methods of Classical Mechanics. 2nd edn. Springer, New York (1989).

\bibitem{ballerin2025so}
F. Ballerin, N. Blaser, E. Grong: $SO(3)$-Equivariant Neural Networks for Learning Vector Fields on Spheres. arXiv preprint arXiv:2503.09456 (2025).

\bibitem{bogatskiy2022equivariant}
A. Bogatskiy, S. Ganguly, T. Kipf, R. Kondor, D. W. Miller, D. Murnane, J. T. Offermann, M. Pettee, P. Shanahan, C. Shimmin, and S. Thais:
Symmetry Group Equivariant Architectures for Physics, 
arXiv preprint arXiv:2203.06153, (2022).

\bibitem{cho2022spherical}
S.~Cho, R.~Jung, and J.~Kwon:
Sampling based spherical transformer for 360 degree image classification, Expert Systems with Applications, vol.~238, 121853, (2024). doi: \href{https://doi.org/10.1016/j.eswa.2023.121853}{10.1016/j.eswa.2023.121853}.

\bibitem{cohen2018spherical}
T.~S.~Cohen, M.~Geiger, J.~K{\"o}hler, and M.~Welling:
Spherical CNNs. Proceedings of the 6th International Conference on Learning Representations (ICLR), (2018). arXiv: \href{https://arxiv.org/abs/1801.10130}{1801.10130}.

\bibitem{CrainicPoisson}
M. Crainic, R. L. Fernandes, I. Mărcuţ: Lectures on Poisson Geometry. Graduate Studies in Mathematics, vol. 217. American Mathematical Society (2021)

\bibitem{dosovitskiy2020image}
A. Dosovitskiy, an others: An image is worth 16x16 words: Transformers for image recognition at scale. Proceedings of the 9th International Conference on Learning Representations (ICLR), (2021). arXiv: \href{https://arxiv.org/abs/2010.11929}{2010.11929}.

\bibitem{poisson}
M.~Á. Evangelista-Alvarado, J.~C. Ruíz-Pantaleón, and P.~Suárez-Serrato:
On computational Poisson geometry I: Symbolic foundations, Journal of Geometric Mechanics, vol.~13, no.~4, pp.~607--628, (2021). doi: \href{https://doi.org/10.3934/jgm.2021018}{10.3934/jgm.2021018}.

\bibitem{numpoisson}
M.~Á. Evangelista-Alvarado, J.~C. Ruíz-Pantaleón, and P.~Suárez-Serrato: 
On computational Poisson geometry II: Numerical methods, 
Journal of Computational Dynamics, vol.~8, no.~3, pp.~273--307, (2021). 
doi: \href{https://doi.org/10.3934/jcd.2021012}{10.3934/jcd.2021012}.

\bibitem{examples}  M.~Á. Evangelista-Alvarado, J.~C. Ruíz-Pantaleón, and P.~Suárez-Serrato: Examples of Symbolic and Numerical Computation in Poisson Geometry , Geometric Science of information GSI (2021),  Lecture Notes in Computer Science , vol. 12829, 200--208.

\bibitem{symflux}
M. Evangelista-Alvarado, P. Suárez-Serrato:  SymFlux: deep symbolic regression of Hamiltonian vector fields. arXiv preprint \href{https://arxiv.org/abs/2507.06342v1}{arXiv:2507.06342} [cs.LG] (2025).


\bibitem{guo2025spatioformer}
Y. Guo, K. Mokany, S.R. Levick, J. Yang, P. Moghadam: Spatioformer: A Geo-encoded Transformer for Large-Scale Plant Species Richness Prediction. IEEE Transactions on Geoscience and Remote Sensing (2025).

\bibitem{hu2025structurepreserving}
J. Hu, J.-P. Ortega, D. Yin: A structure-preserving kernel method for learning Hamiltonian systems. arXiv preprint arXiv:2403.10070 (2025).

\bibitem{Koszul2019}
J.L. Koszul, Y.M. Zou: Introduction to Symplectic Geometry. Springer, Singapore (2019).

\bibitem{lai2023spherical}
X. Lai, Y. Chen, F. Lu, J. Liu, J. Jia: Spherical transformer for lidar-based 3d recognition. In: Proceedings of the IEEE/CVF Conference on Computer Vision and Pattern Recognition, pp. 17545--17555 (2023).

\bibitem{geomstats}
N.~Miolane, N.~Guigui, A.~Le Brigant, J.~Mathe, B.~Hou, Y.~Thanwerdas, S.~Heyder, O.~Peltre, N.~Koep, H.~Zaatiti, H.~Hajri, Y.~Cabanes, T.~Gerald, P.~Chauchat, C.~Shewmake, D.~Brooks, B.~Kainz, C.~Donnat, S.~Holmes, and X.~Pennec:  
Geomstats: A Python Package for Riemannian Geometry in Machine Learning, 
\textit{Journal of Machine Learning Research}, vol.~21, no.~223, pp.~1--9, (2020). 

\bibitem{russwurm2024spherical}
M. Rußwurm, K. Klemmer, E. Rolf, R. Zbinden, and D. Tuia: Geographic location encoding with spherical harmonics and sinusoidal representation networks, in Proceedings of the Twelfth International Conference on Learning Representations, (2024).

\bibitem{vaswani2017attention}
A. Vaswani, N. Shazeer, N. Parmar, J. Uszkoreit, L. Jones, A.N. Gomez, {\L} Kaiser, I. Polosukhin: Attention is all you need. Advances in neural information processing systems, 30 (2017).

\end{thebibliography}
\end{document}